\documentclass{article}
\pdfoutput=1
\usepackage{arxiv}
\usepackage[utf8]{inputenc} 
\usepackage[T1]{fontenc} 
\usepackage{hyperref}  
\usepackage{url}   
\usepackage{booktabs}  
\usepackage{amsfonts}  
\usepackage{nicefrac}  
\usepackage{microtype}  
\usepackage{lipsum}
\usepackage{multicol}
\usepackage[labelformat=empty]{caption}
\usepackage{graphicx}
\graphicspath{ {./images/} }
\usepackage{float}
\usepackage{amsmath} 
\usepackage{enumitem}

\title{Testing Components of the Attention Schema Theory in Artificial Neural Networks}

\author{
 Kathryn T. Farrell$^1$, Kirsten Ziman$^1$, Michael S. A. Graziano$^{1,2}$\\\\
 $^1$Princeton Neuroscience Institute\\
 $^2$Department of Psychology\\
 Princeton, NJ 08544 
 }

\begin{document}
\maketitle
\begin{abstract}
Growing evidence suggests that the brain uses an attention schema, or a simplified model of attention, to help control what it attends to. One proposed benefit of this model is to allow agents to model the attention states of other agents, and thus predict and interact with other agents. The effects of an attention schema may be examined in artificial agents. Although attention mechanisms in artificial agents are different from in biological brains, there may be some principles in common. In both cases, select features or representations are emphasized for better performance. Here, using neural networks with transformer attention mechanisms, we asked whether the addition of an attention schema affected the ability of agents to make judgements about and cooperate with each other. First, we found that an agent with an attention schema is better at categorizing the attention states of other agents (higher accuracy). Second, an agent with an attention schema develops a pattern of attention that is easier for other agents to categorize. Third, in a joint task where two agents must predict each other to paint a scene together, adding an attention schema improves performance. Finally, the performance improvements are not caused by a general increase in network complexity. Instead, improvement is specific to tasks involving judging, categorizing, or predicting the attention of other agents. These results support the hypothesis that an attention schema has computational properties beneficial to mutual interpretability and interactive behavior. We speculate that the same principles might pertain to biological attention and attention schemas in people.
\end{abstract}

\section*{Introduction}
\begin{multicols}{2}

\par Growing evidence suggests that the brain uses an attention schema, or a simplified model of attention, to monitor, predict, and help control attention [1-5]. The attention schema has also been hypothesized to play a prominent role in social cognition. It may be used to model the attention of others, affording better theory of mind, better predictions of the behavior of others, and therefore more effective social interaction [1,6-12]. One way to test hypotheses about a proposed computational mechanism in the brain, such as an attention schema, is to incorporate it into artificial agents, where its consequences can be tested in a controlled manner. In the present study, we tested whether adding an attention schema to an artificial agent improves its ability to make judgements about the attention of other agents, improves its ability to have its own attention judged by others, and improves its ability to coordinate with another agent in a joint task. 

\par The biological mechanism of selective attention in the human brain, at least as it is understood, may be quite different from the kinds of attention that have been implemented in artificial agents. A great variety of attention mechanisms have been tested in machine learning, some more inspired by biological attention and some less. One particular method, associated with transformer architecture [13], has become extremely successful at helping artificial agents learn complex tasks, and now lies at the root of most large artificial intelligence (AI) agents such as modern chatbot technology. The extent of similarity or difference between transformer attention and biological, human attention has been controversial. Original accounts described the mechanism as similar, but later interpretations suggested that fundamental differences exist [14]. In the present study, we were inspired by proposed principles of the attention schema as derived from human cognitive neuroscience [1-12], and yet we tested these principles in an artificial transformer architecture. There are several reasons for this choice. First, we wished to examine the utility of the biologically-inspired attention schema when applied to the most common and successful mechanism of attention currently used in artificial systems. Though versions of attention schemas have been tested using other kinds of artificial attention [15-17], these previous studies have had limited impact and generated limited interest because they do not involve attention mechanisms that are mainstream in machine learning, and therefore have limited practical applicability.

\par Second, and most importantly, the specific mechanism by which attention is implemented should not change the underlying principles at work. Although the mechanisms may be quite different between biological and transformer attention, the deeper principles do share similarities, and it is these principles that matter most. Selective attention, whether in a human brain or an artificial transformer architecture, is fundamentally about selecting some features to have a greater influence over downstream information processing and therefore, ultimately, the behavior of the agent. The central concept in the attention schema theory is that, because of the importance of this selection, attention is one of the most important factors shaping behavior. Therefore, to better predict behavior, it is useful to build a predictive model of attention; and therefore, there is utility in building a predictive model of one’s own attention, and a social utility in building a model of the attentional states of others. This underlying concept is not tied to the specific implementation details of attention. For example, whether attention emerges from a biased competition mechanism of neuronal inhibition and excitation involving top-down and bottom-up signals, as for example some have theorized for the human brain, or whether attention emerges as a result of a transformer architecture involving multiple attention heads, as is often implemented in machine learning, the deeper concepts remain. A predictive model of an agent’s own attention mechanism, and of the attention mechanisms of other agents, should be of benefit, especially in social tasks that require fine-tuned prediction of the behavior of other agents. If these principles, inspired by the biological case, do apply to the very different implementation of transformer attention, then they may prove to be of great utility in building effective AI, and most especially, in shaping agents that have better social cooperative abilities. 

\par Recently, versions of attention schemas have been incorporated into artificial neural networks to test whether the same principles of the attention schema inspired by data from humans might provide benefits in machine learning, and to determine whether lessons from the more controlled, artificial case might provide conceptual insights into human cognition [15-18]. Several studies suggest that an agent that controls movement of a spotlight of visual attention can do so more effectively when it includes a version of an attention schema to monitor and predict its attentional state [15-17]. These studies were the first to show the validity of the attention schema hypothesis in an artificial, computational context. However, they addressed only whether an attention schema aids an agent in controlling its own attention, while leaving untested the much more complex question of whether an attention schema aids an agent in making inferences about the attentional states of others or in the performance of interactive tasks with others. 

\par One recent study suggests that an attention schema may improve performance of artificial systems in cooperative tasks. Liu et al. [18] trained interacting groups of networks on visual tasks that required the agents to coordinate with each other. In that study, a transformer-style visual attention was used. In some conditions, networks were provided with an attention schema, in the form of a learned ability to predict the state of their own attention mechanism. With an attention schema, agents performed the social coordination tasks better. One interpretation is that when an agent is trained to predict its own attention states, it also becomes better at predicting the attention states of other agents in the same environment. In that hypothesis, learning to self-model leads to an improvement in the ability to model others and thus to an improvement in group task performance. 

\par A second recent study reported unexpected consequences for agents that build predictive self-models. An attention schema is a specific type of self-model, since it is a predictive model of the attentional state of an agent, but one can study the general impact on an agent of learning to predictively model any aspect of itself. Premakurma et al. [19] trained artificial networks on a variety of primary discrimination tasks, including both visual and linguistic tasks. In some conditions (the self-model conditions), in addition to the primary task, the agents were also trained on an auxiliary task in which they were rewarded for predicting the activation states of their own inner layers. The networks that were trained on the self-modeling task showed consistent changes. They learned to make themselves simpler, more regularized, and more parameter-efficient. These gains were found across all types of networks studied. It was hypothesized that when a network is given the auxiliary task of predicting its own internal activity states, it learns to make those internal states simpler and more regular. The simpler the internal states, the easier they are to predict. Therefore, as the network learns to predict itself, it learns to simplify itself. According to that hypothesis, learning a self-model restructures the self, regularizing internal processes, and making internal states an easier target of predictive modeling. 

\par Based on these prior findings, we proposed the following two hypotheses about the relationship between self-models and social behavior in artificial agents. First, when an agent learns to self-model, such as when it learns an attention schema that models its own attention state, it can transfer some of that learning toward predictively modeling other agents. Predictive modeling of other agents then becomes a useful part of interacting with them. Second, when an agent learns to self-model, it restructures itself to be more regularized and more predictable, and therefore the agent becomes a more effective target for being predictively modeled by other agents. Both changes – becoming better at modeling others, and becoming a more transparent target for being modeled by others – should render the agent a more effective member of a cooperative group. 

\par The purpose of the present study was to test these hypotheses about how attention schemas might influence multi-agent interaction in artificial agents using a transformer style of attention. We conducted three related experiments. Experiment 1 asked: Does an agent that has an attention schema that models its own attention also have an improved ability to model the attention states of other agents? Experiment 2 tested an alternative explanation: Are the performance benefits of an attention schema merely a non-specific effect of added network complexity? Finally, Experiment 3 addressed the question of multi-agent cooperation: Do agents with attention schemas perform better on a simple joint task?

\end{multicols}

\section*{Experiment 1: Do attention schemas improve the ability of agents to model each other’s attention states?}
\label{sec:headings}
\subsection*{Introduction to Experiment 1}
\begin{multicols}{2}
\par The purpose of experiment 1 was to train artificial agents with and without attention schemas. The agents used a transformer type of attention. The attention schema, or learned predictive model of its own attention state, was adapted to the transformer type of attention mechanism, as described below. We tested two hypotheses. Hypothesis 1 was that when an agent has an attention schema, it will be better at learning to make categorical judgements about the attention of other agents. Hypothesis 2 was that when an agent has an attention schema, other agents will be better at learning to make categorical judgements about its attention.

\par Agents were first trained on a visual categorization task (e.g. categorize a picture as showing a golf ball or a garbage truck). We then froze the weights in all layers except a final block, and trained the agents on a new task, the “attention judgement” task. The task allowed us to test the ability of one agent to make categorical determinations about the attention states of another agent. The attention judgment task was modified from Ziman et al. [20], who demonstrated the ability of people to assess the attention states of others. In the human version of the task, participants looked at a spotlight of attention moving around a picture and judged whether the spotlight represented real, overt attention (based on eye movements recorded previously from a real person), or whether it represented a randomly scrambled version of attention (the recorded attention locations presented in a random order). Since people were able to perform above chance, the human brain must contain information about the typical or predicted patterns of attention and must be able to recognize when those patterns have been violated. Thus, humans demonstrate an implicit predictive model of attention. It is not simply the case that they can learn to distinguish one arbitrary type of pattern from another – rather, they know, without any feedback, which patterns are typical of real human attention and which ones are not.

\par To modify the task for the present study such that it could be applied to a transformer type of attention, we assigned agents to one of two roles: either the receiver or the sender. If the agent was assigned the sender role, it performed its originally trained visual categorization task. The attention activation values, for each trial of the task, were saved in the form of a multidimensional tensor. In one version, the tensor was left in its correct, original form. In another version, the attention values were randomly scrambled along one dimension of the tensor. 

\par If the agent was assigned the receiver role, then it was first trained on the image categorization task; then its weights were frozen except in the final block of layers; and then it was trained on the attention judgement task. On each trial of the attention judgement task, instead of receiving a picture as an input, the agent was provided with an attention tensor; and instead of being rewarded for categorizing the picture, it was rewarded for categorizing the tensor as either a valid example of attention (recorded from the performance of a sender agent) or as an invalid example (constructed by randomly modifying a valid tensor). The task therefore tested the ability of the receiver agent to learn to recognize violations of predictable patterns in the attention behavior of the sender. 

\par The experiment used a 2$\times$2 design: the sender agent could be trained with or without an attention schema (a learned ability to predict its own attention states), and the receiver agent could be trained with or without an attention schema. We proposed the following hypotheses. 

\par \textit{Hypothesis 1:} If the receiver agent was pre-trained with an attention schema, then it will be better able to learn the attention judgement task. Such a result would suggest that once the agent is trained to use an attention schema to model its own attention, then it becomes better at learning to model the attention behavior of other agents. 

\par \textit{Hypothesis 2:} If the sender agent was trained with an attention schema, then the receiver agent will be able to learn the attention judgement task better. Such a result would suggest that when an agent contains an attention schema, it is, itself, an easier target for other agents to build predictive models of it. It would suggest that having an attention schema adjusts or restructures the agent’s own attention, to make it more easily interpretable to others. 
 
\subsection*{Methods for Experiment 1}
\par We trained networks on a visual categorization task. Given an image as input (for example, a picture of a golf ball on a lawn), the network was trained to arrive at a dichotomous categorization (for example, golf ball versus garbage truck). All networks and task environments were created using PyTorch [21] and all experiment code and data are publicly available at \url{github.com/kathrynfarrell/attention_schemas_in_anns}. We used an image-adapted implementation of the Attention Schema Neural Network architecture proposed by Liu et al. [18], which utilized a transformer-based attention mechanism. Figure 1 shows the architecture, with 1A showing the control architecture (control network: attention but no attention schema) and 1B showing the architecture that includes an added attention schema (attention schema network). It is important to note that the task of categorizing one type of picture from another is not necessarily attentionally demanding for humans, but relies on the transformer attention mechanism in the artificial agent. To solve the task, the agent must be able to select features in the picture to have greater impact, and it is the transformer encoder that performs that function. Thus, this is an appropriate task for testing the attention hypotheses as applied to these artificial agents.

\begin{figure*}
 \centering
 \includegraphics[width = 14 cm]{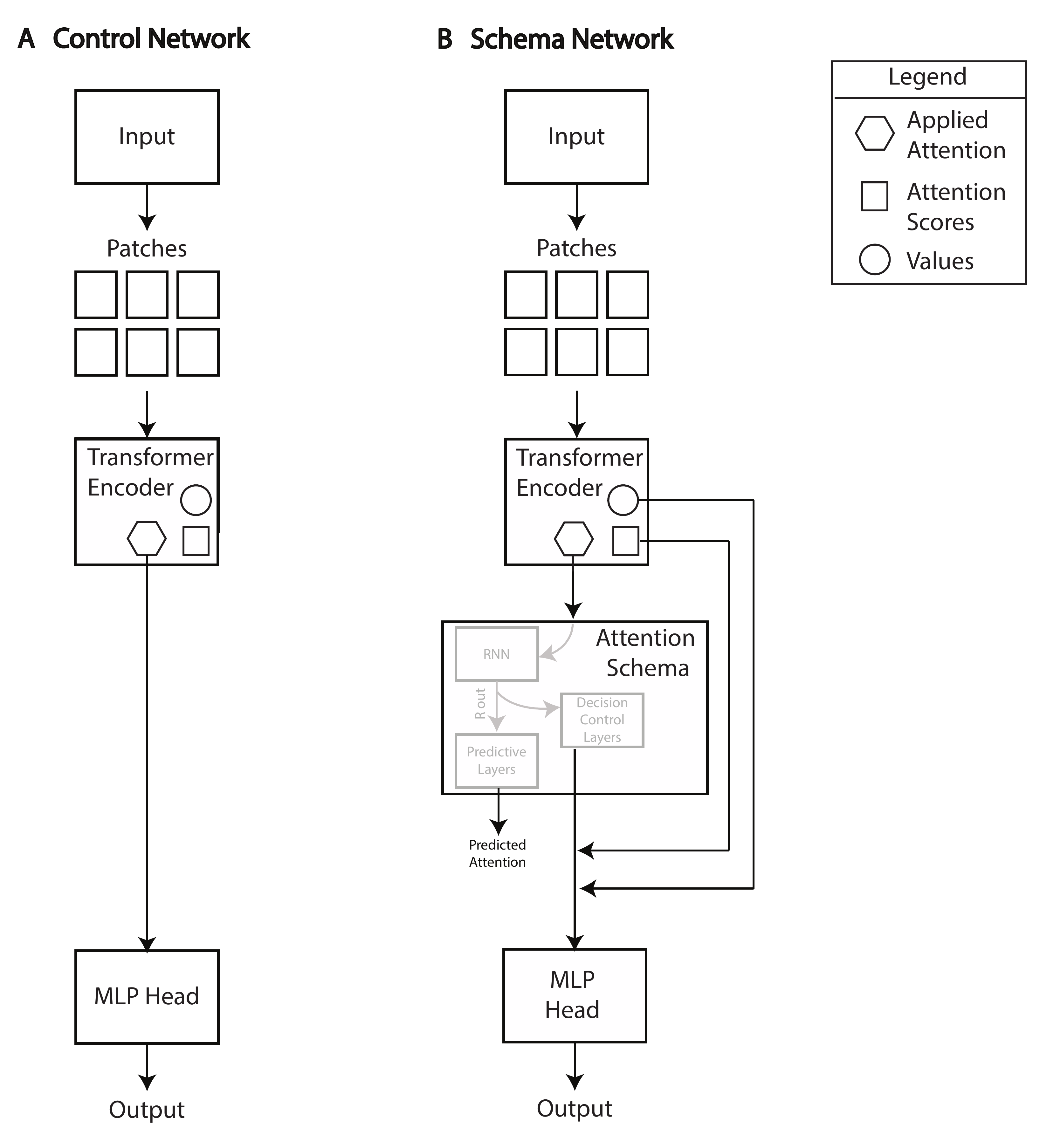}
 \caption{
   \textit{Figure 1: The Attention-Schematic Neural Network (ASNN) architecture.} A. The control model without an attention schema. B. The model with an attention schema, modified from Liu et al. [18]. The model is based around a standard vision transformer (ViT) block (“Transformer Encoder”), which generates attention scores and applies them to the network’s representation of the image (Values), to produce a representation of the image with the attention applied (“applied attention”). Conceptually, the attention scores can be thought of as a heatmap of where to focus and the Applied Attention can be thought of as the heatmap overlaid onto the image. The applied attention is then passed into a Recurrent Neural Network (RNN) module with learnable parameters. The output of this module (“$R_{out}$”) is a precursor to both the network’s final attention prediction and its dynamic selection of attention scores, both of which are produced by several linear layers with learnable parameters (the Predictive Layers and the Decision Control Layers, respectively). For the attention-revision computations, attention suppression or activation is achieved through processes beginning in the Decision Control Layers. Linear “suppressor” and “activator” layers in the Decision Control produce logit pairs, which are binarized using a Gumbel-Softmax distribution, then combined via element-wise multiplication. The binary mask resulting from that multiplication is then applied to the original attention scores (by element-wise multiplication) yielding the final attention—a sparser version of the original attention scores where some of the scores have been selectively “shut off” or reduced to zero. The final attention is then applied to the original image representation (Values), and the result is passed to the final MLP head, which classifies the image as belonging to a particular category. This MLP head is the only part of the network that continues updating during transfer learning (specifically, the final multi-layer perceptron in the MLP head), while the rest of the network remains frozen.
}
\end{figure*}

\par For details on how the attention schema mechanism works, see Liu et al. [18]. Briefly, the objectives of the network with the schema are twofold: learning to model and refine its own attention and learning to complete a task (e.g. classifying images). The schema network uses its applied attention in service of both goals: modeling its own attention (by predicting its final attention), and further editing the attention to its final refined state. To accomplish this, the schema network uses a transformer block with six attention heads, and passes the applied attention into a recurrent neural network module with learnable parameters. The use of multiple attention heads is standard in transformer architectures. The duplication allows for better performance. Again, although this architecture is very different from the implementation of attention in the human brain, it is the most effective way to test whether the general principles of an attention schema apply to standard implementations in machine learning. The output of this module ($R_{out}$) is further processed and used to model attention; specifically, it is passed into the Predictive Layers, which output the predicted attention. $R_{out}$ is also further processed and used to refine attention; it is passed into the Decision Control Layers which produce the final refined attention (with the Decision Control Layers employing activator units, suppressor units, and categorical reparameterization with gumbel-softmax). In this way, manipulations of the initial applied attention are used to generate both the prediction of the final attention as well as the final attention itself. Schema networks are trained on their combined ability to accurately complete the main task (e.g. classifying an image) and predict their own attention.

\par In the present study, the network received as input an image of size 256 pixels $\times$ 256 pixels $\times$ 3 color channels, and was trained to produce an output that classified the image into one of two categories. To ensure that the result was not specific to one type of categorization, separate networks were trained on three different sets of images and categorizations (image classification task A, parachute versus bench; image classification task B, French horn versus cassette player; image classification task C, golf ball versus garbage truck). All three image sets were subsampled from Imagenette [22]. Two kinds of networks were trained: control networks without an attention schema, and networks with an attention schema. For those with an attention schema, the relative weight on the main task versus the attention prediction task was 0.95 to 0.05.

\par Each network was trained on 1,912 images for 20 training epochs, and tested on 730 images. For each network that did not contain an attention schema, for each of its 730 testing discriminations, we saved the attention scores from three of its six attention heads (randomly selected) in the transformer encoder block. For each network that contained an attention schema, we used a similar procedure but saved the modified attention scores that resulted from element-wise multiplication between the attention scores and the output of the decision control layers. 

\par To produce an attention tensor, we stacked the scores from the three attention heads into a 257 token $\times$ 257 token $\times$ 3 attention head tensor (257 tokens in the first two dimensions resulting from the 16 $\times$ 16 image patches from the original input image plus one class token). From these tensors, we created the veridical and scrambled attention data that would be sent to the receiving networks for discrimination. The veridical attention tensors were created by reducing the size of the tensors from 257 $\times$ 257 $\times$ 3 to 256 $\times$ 256 $\times$ 3, matching the size of the original input image. For the scrambled attention, we randomly permuted each tensor along its last dimension, which randomly shuffled the attention scores within the data from each attention head. Then, we reduced the resulting tensor to 256 $\times$ 256 $\times$ 3, again matching the dimensions of the original input image. This procedure resulted in a set of 1460 attention tensors from each network, half veridical and half scrambled, with the same dimensions as the input images the networks were originally trained on. 

\par Receiving networks were constructed by first pre-training them on the image discrimination task following the same procedures described above. We then froze all weights in the network except those in the final multilayer perceptron (MLP) head (a module consisting of two linear layers, which computes the final classification decision). We further trained the network for 800 training epochs by providing the network with new training data in the form of 1,314 labeled attention tensors saved from the sending networks, half veridical and half scrambled. The receiving networks were then tested on an additional 146 attention inputs (half veridical, half scrambled). To avoid any possible bias from the original image classification task, receiving networks were pre-trained on one of the three image classifications only (e.g., classification A) and were then trained to judge attention data from networks trained on a different image classification task (e.g., B or C). At the end of training, the performance accuracy of the receiving network was saved. Since three receiving networks were trained, one for each of the three classification tasks, three accuracy scores were obtained. We then repeated the process 9 times, resulting in 27 accuracy scores. For all results displayed, each point represents a mean of 27 measurements. 

\subsection*{Results for Experiment 1}

\par Figure 2 shows the results for experiment 1. Performance on the attention discrimination task was well above chance (mean = 80.87\% correct, SEM = 2.38) when both the sending network and the receiving network contained an attention schema. Performance dropped when the sending network contained an attention schema but the receiving network did not (mean = 70.61, SEM = 2.37). Performance dropped further when the sending network did not have an attention schema, but the receiving network did (mean = 54.58, SEM = 1.25). Finally, performance was lowest when neither sending nor receiving network contained an attention schema (mean = 52.78, SEM = 1.21). These results support both hypotheses. With respect to hypothesis 1: When the receiving network contained an attention schema, it had an advantage in learning to categorize the attention data from the sending network. With respect to hypothesis 2: When the sending network contained an attention schema, the network was in some manner altered such that its attention data was easier for the receiving network to categorize. 

\begin{figure}[H]
 \centering
 \includegraphics[width = 8 cm]{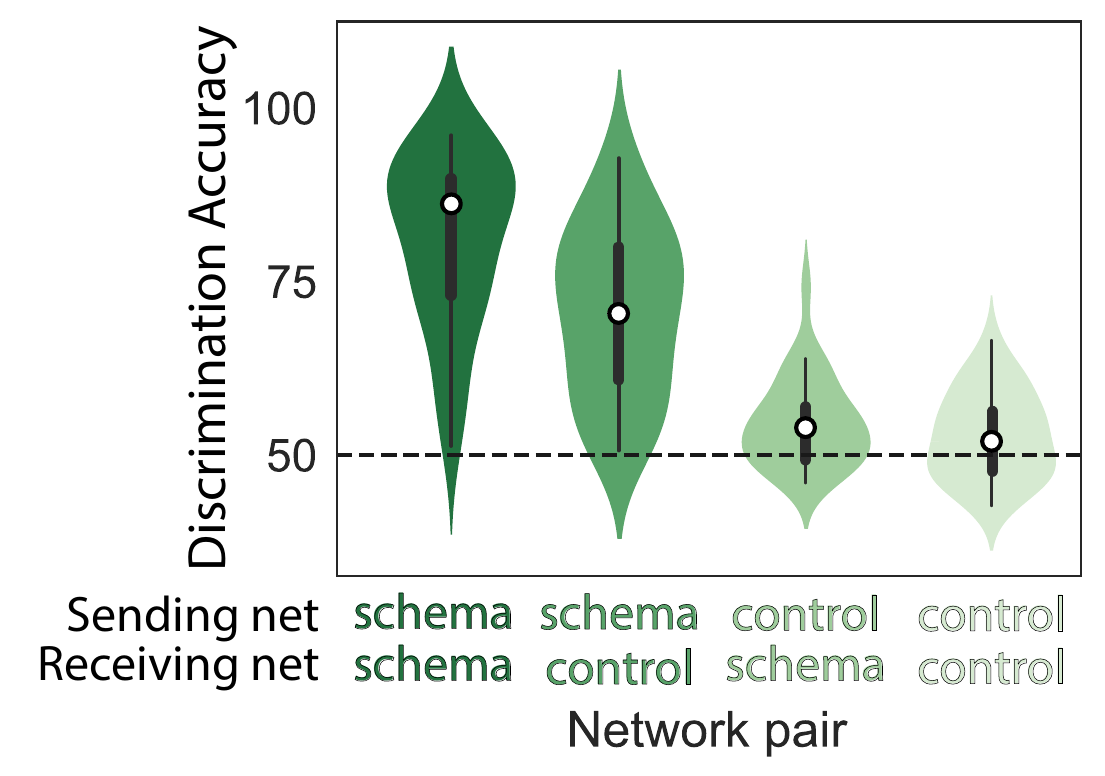}
 \caption{
   \textit{Figure 2: Results for Experiment 1.} Accuracy on the attention discrimination task as a function of whether the sending network had an attention schema or was a control network, and whether the receiving network had an attention schema or was a control network. Each violin plot represents 27 data points. The violin shape represents a kernel density estimate of the distribution. Inside each violin is a boxplot, with a white dot representing the median, thick black bars spanning the inter-quartile range, and thin black bars spanning the 1.5 inter-quartile range.
}
\end{figure}

\par Figure 2 shows the result. Performance on the attention discrimination task was well above chance (mean = 80.87\% correct, SEM = 2.38) when both the sending network and the receiving network contained an attention schema. Performance dropped when the sending network contained an attention schema but the receiving network did not (mean = 70.61, SEM = 2.37). Performance dropped further when the sending network did not have an attention schema, but the receiving network did (mean = 54.58, SEM = 1.25). Finally, performance was lowest when neither sending nor receiving network contained an attention schema (mean = 52.78, SEM = 1.21). These results support both hypotheses. With respect to Hypothesis 1: When the receiving network contained an attention schema, it had an advantage in learning to categorize the attention data from the sending network. With respect to Hypothesis 2: When the sending network contained an attention schema, the network was in some manner altered such that its attention data was easier for the receiving network to categorize. 

\par These effects were statistically significant. We performed a 2 $\times$ 2 ANOVA and found that the main effect of the receiving network (receiving network does or does not have an attention schema) was statistically significant (F=10.17, p=0.0018, $\eta^2$p=0.09, with a post-hoc power analysis indicating statistical power of 0.77). The main effect of the sending network (sending network does or does not have an attention schema) was also statistically significant (F=136.18, p=1.28 $*$ 10$^{-20}$, $\eta^2$p=0.57). The interaction between the two factors was also significant (F=4.99, p=0.027, $\eta^2$p=0.05). The largest effect was whether the sending network contained an attention schema. Having an attention schema made the network more interpretable to other networks.

\begin{figure}[H]
 \centering
 \includegraphics[width = 8 cm]{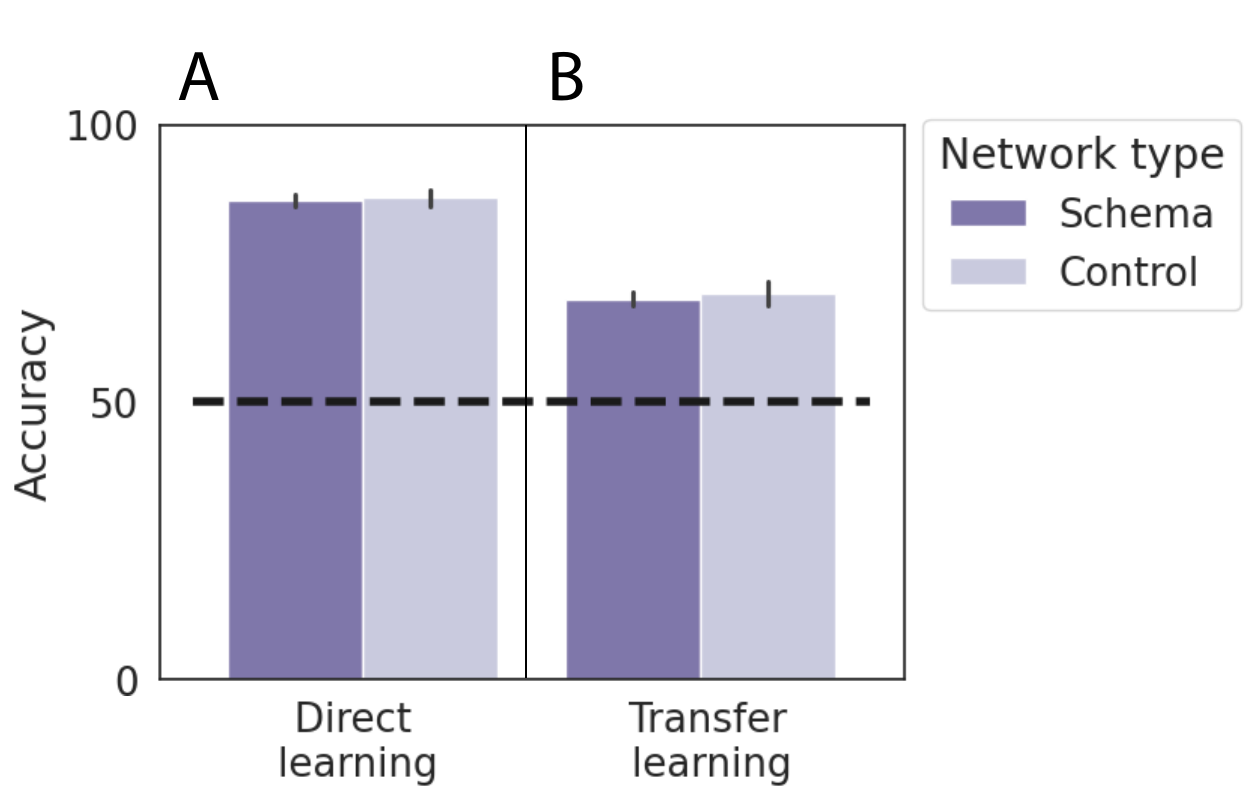}
 \caption{
   \textit{Figure 3. Performance of Schema and Control networks on image classification tasks.} A. Mean accuracy on image classification task for schema versus control networks, in experiment 1. Bars show 95\% confidence interval. No significance difference was found. B. Results for experiment 2. Mean accuracy of schema versus control networks on transfer learning (pretraining on one image classification task followed by transfer learning to a second image classification task). Bars show 95\% confidence interval. No significance difference was found.
}
\end{figure}

\par We also examined performance of the networks on the visual categorization task. We did not have any specific hypothesis about whether a network with an attention schema would perform the visual categorization task better or worse than a network without an attention schema. One possibility is that the attention schema would improve the agent’s ability to refine its own attention, and thus improve its performance on the categorization task. Another possibility is that the presence of the attention schema would provide benefits only for the more “social” aspects of performance, categorizing the attentional states of other agents. Figure 3A shows the result. When analyzing all networks trained on the visual categorization task, comparing those with and without an attention schema, we saw no significant difference in performance (t=-0.34, p=0.73, d=0.09, post-hoc power=0.06). While acknowledging that null results must be treated with caution, we interpret this result as suggesting that adding an attention schema to a network does not make the network automatically perform better whether due to more refined attention or due to a non-specific effect of adding computational complexity. This question of the specificity of the effect is revisited in Experiment 2. From the data thus far, it appears that the addition of the attention schema improved performance in a specific manner: it improved the ability of one network to judge the attention states of another network, and it improved the ability of one network to produce attention states that are more easily judged by another. In this sense, it improved more “social” aspects of performance.

\subsection*{Discussion for Experiment 1}

\par Experiment 1 showed that the presence of an attention schema in a network (the trained ability to predict its own attention states) confers two specific advantages, at least in one kind of network with a specific implementation of attention and of an attention schema. The first advantage is that when a network with an attention schema is pretrained on a task, that network is then better able to learn to make judgements about the attention states of other networks. Through initial training in which it predicts its own attention states, it gains an improved ability to model the attention states of others. The second advantage is that when a network contains an attention schema, its attention states become easier for other networks to interpret and categorize. In principle, both of these changes should be advantageous in joint tasks in which interpreting, categorizing, and predicting the behavior of others is useful. It may be of interest that the largest impact we observed here was that when a network contained an attention schema, its attention states were more easily recognized and categorized by other agents. It has been suggested that an important feature of social interaction, whether in people or in simple artificial agents, is that self-models (such as attention schemas) help to restructure networks, causing them to make their internal computations simpler and more regularized, thus making those networks an easier target for others to predict and interpret [19]. The present results are consistent with that hypothesis.
\end{multicols}

\section*{Experiment 2: Are the performance benefits of an attention schema caused by added network complexity?}

\begin{multicols}{2}

\par In Experiment 1, attention schemas improved the ability of networks to categorize each other’s attention states (showing improved performance on the attention categorization task), but did not improve the ability of networks to categorize visual images in general (showing no significant change on the image classification task). This specificity of the effect suggests that the benefit of an attention schema is not just the general result of added network complexity, but instead a specific improvement in the ability to learn about attention patterns. 

\par However, the results of Experiment 1 are subject to an alternative hypothesis. The image classification task was learned by networks first. After being trained on that initial task, networks had their weights frozen except in the final MLP block, and then underwent transfer learning on the attention categorization task. Could it be that having an attention schema does not benefit performance on a primary task, but adds enough extra computational complexity to benefit performance on transfer learning to a secondary task? Would transfer learning to any secondary task be improved by an attention schema, or is the improvement specific to the attention categorization task?

\par To check this possibility, we examined the networks’ performance in transfer learning. First, we pretrained them on the image classification task, replicating the procedure in experiment 1. However, instead of then training the networks to classify attention (real versus shuffled), we trained the networks on another image classification task. Networks that were pretrained on one of the three image classification tasks (for example, task A) were retrained on a different one (for example, task B). We then asked whether transfer learning from one classification task to another was better for networks that contained an attention schema. 

\par The results are shown in Figure 3B. Transfer learning was comparable between the two network architectures. No significant difference was found (t=-0.74, p=0.46, d=0.20, post-hoc power=0.111). While null results should be considered with caution, these results generally suggest to us that schema networks and control networks perform similarly on basic discrimination tasks both before and after transfer learning. 

\par The results of Experiment 2 help to confirm that the effect of an attention schema, observed in Experiment 1, was not a nonspecific result of greater network complexity. The improvements in performance seen in experiment 1 were specific to categorizing attention performance, and did not extend to image categorization in general.

\end{multicols}
\newpage
\section*{Experiment 3: Do agents with attention schemas perform better on a cooperative task?}
\subsection*{Introduction to Experiment 3}

\begin{multicols}{2}

\par A central hypothesis of the attention schema theory is that, in people, the presence of an attention schema allows for better social cognition and better performance in cooperative tasks [1,6]. The reason is that, at a simple, fundamental level, attention influences behavior, and therefore predicting attention state allows for better prediction of behavior. The purpose of the present set of studies was to examine how well the principles of attention schemas apply to a standard transformer-type artificial neural network. In experiment 1, networks that contained attention schemas showed better ability to categorize each other’s attention states. But are these networks actually better able to cooperate, if tested directly on a joint task? We hypothesized that networks with attention schemas would perform better on a joint task that requires interaction with another agent. 

\par We chose a simple task involving two networks mutually coloring an image, much like two children sharing a page in a coloring book. The networks took turns placing colored pixels on the image. Reward was earned jointly by both networks for painting as many pixels as possible over multiple turns, but reward was decreased if the painted area of one network overlapped the painted area of the other network. On each turn, each agent chose which pixels to color. While each agent had information about its partner’s behavior on previous turns, it had no information about which pixels its partner would choose on the current turn. Ideal performance on the task should require each network to make turn-by-turn predictions of the behavior of its partner. If one network could predict the part of the image that its partner was likely to color, it could then choose to color a different part of the image, avoiding overlap. Thus, mutual behavioral prediction should improve performance. The behavior of each network was in turn influenced by its visual attention to the underlying image. We therefore reasoned that it would be beneficial for each network to learn an implicit, predictive model of the other network’s attention. Although a mutual coloring task, because of its simplicity, may not be an ideal test of social cooperation or mutual attention modeling in humans, it seemed, for the reasons just explained, to be well suited to testing the hypothesis in the artificial networks studied here.

\subsection*{Methods for Experiment 3}

\par Using an image-based Multi-Agent Reinforcement Learning (MARL) environment, we designed a coloring task wherein, over the course of multiple turns, two networks colored in an image. The two networks were presented with the same image selected from a pool of 2,000 images, subsampled from 10 categories in Imagenette [22]. On each turn, each network chose a selection of pixels from the image and marked them as painted. The network could see which pixels had previously been painted and which member of the team had painted it, but had no information about which pixels its partner network would choose to color on the current turn. Reward, given jointly to both networks, increased with each novel pixel that was colored on each turn, but decreased for each pixel that was colored by both networks on that turn (“overlapping” pixels). The reward was calculated according to the formula:
\begin{center}
  \scalebox{1.2}{
  $R = \frac{\alpha_{\text{discovery}}*n_{\text{discovery}}}{1+\alpha_{\text{overlap}}*n_{\text{overlap}}}$
  }
\end{center}

where $n_{\text{discovery}}$ denotes the total number of novel pixels selected by both agents on a given turn, $n_{\text{overlap}}$ denotes the number of overlapping pixels selected by both agents on that turn, and $\alpha_{\text{discovery}}$ and $\alpha_{\text{overlap}}$ signify scaling coefficients for each of these quantities (0.5 and 1.7, respectively). Each coloring episode ended either after the networks had colored every pixel in the image, or after the same number of turns had passed as there were pixels in the image (whichever occurred first). 

\par The networks we used for this task had the same underlying structure as those in Experiment 1 but were modified to do an image coloring task instead of a classification task. The training method was also modified to use reinforcement learning. For more details on the networks, environments, and training methods in Experiment 3, please refer to the publicly available code (\url{github.com/kathrynfarrell/attention_schemas_in_anns}). 

\par Each network in the interacting pair could have an attention schema or could lack one, resulting in three types of image-coloring teams: schema-schema teams, mixed teams (where one network had an attention schema and one did not), and control-control teams (where neither network had an attention schema). We predicted that schema-schema teams would show the best joint performance, achieving the highest rewards, and control-control teams would show the poorest performance, obtaining the lowest rewards. For each type of pairing, we trained the networks for 50 epochs, 6,000 game turns per epoch. We then repeated each test 15 times. 

\subsection*{Results for Experiment 3}

Figure 4A shows performance on the task across fifty training epochs, for all three types of network parings. In line with our predictions, we found that schema-schema teams achieved the highest rewards on the cooperative coloring task (blue line in Figure 4A, averaged across epochs, mean=2.04). Mixed teams received the second highest rewards (green line in Figure 4A, averaged across epochs, mean=1.91). Control-control teams obtained the smallest rewards (orange line in Figure 4A, averaged across epochs, mean=1.76).

\begin{figure*}
 \centering
 \includegraphics[width = 18 cm]{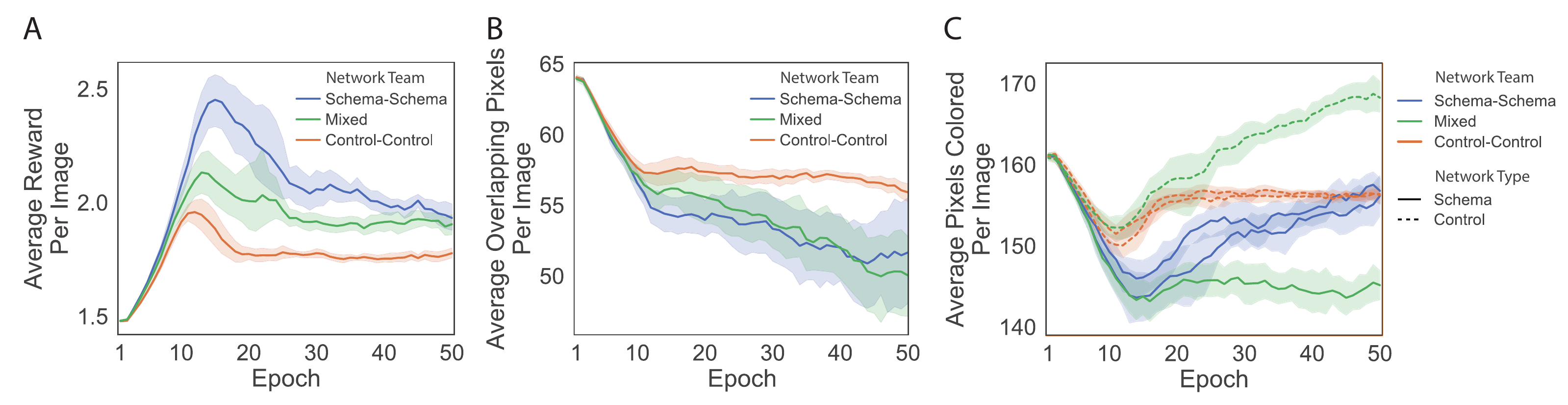}
 \caption{
   \textit{Figure 4: Results for Experiment 3.}  Performance of network pairs on joint coloring task. A. Average reward accrued during the coloring of each image, as a function of training epoch. Shading shows 95\% confidence interval. B. Average amount of overlapping pixels for each image, as a function of training epoch. Shading shows 95\% confidence interval. C. Average number of pixels colored per image, for each network in the pair, as a function of training epoch. For the schema-schema pair, both networks had an attention schema. For the control-control pair, neither network had an attention schema. For the mixed pair, one network had an attention schema and the other did not. Shading shows 95\% confidence interval.
}
\end{figure*}

\par We also analyzed the average number of overlapping pixels as a metric of within-team coordination, since networks that are well coordinated with their teammates should be better at reducing the overlap. The results are shown in Figure 4B. We found that schema-schema teams and mixed teams exhibited better coordination, selecting fewer overlapping pixels on average (blue line for schema-schema teams, mean across epochs=54.51; green line for mixed teams, mean across epochs=54.89), relative to the control-control teams (orange line for control-control teams, mean across epochs=57.79). 

\par By the rules of the game, both networks in a team earned the same number of points, and both networks accrued the same number of overlapped pixels. However, each network chose a different number of pixels to color on each trial, and we analyzed that number. The two blue lines in Figure 4C show the results for the schema-schema team. Across training epochs, the two networks chose a similar number of pixels to color. The average number of pixels selected by the networks first dropped, probably because the networks learned that choosing a large number of pixels on each trial resulted in a large amount of overlap with its partner’s choices. However, after epoch 15, the average number of pixels selected by networks began to rise. At the same time, the average overlap in their selections decreased, as shown in Figure 4B. This pattern suggests that the networks in the schema-schema team gradually learned to make bigger selections, as they became better at coordinating with each other and avoiding overlap.

\par The two orange lines in Figure 4C show the result for the control-control team. Once again, the two networks in the team chose a similar number of pixels to color. Around epoch fifteen, the curve plateaus, suggesting that the networks had learned a useful number of pixels to paint on each turn – enough to garner points, but not so many as to risk too much overlap. 

\par The two green lines in Figure 4C show the result for the mixed, control-schema team. Here the behavior is markedly different. The solid green line shows the trajectory for the network that contained an attention schema. The average number of pixels selected by the network first dropped, then stayed low for the remainder of the epochs. This behavior suggests that of the two networks in the team, the network with an attention schema was less well able to predict where its partner would place colored pixels. As a result, its best strategy was to learn to place relatively few pixels on each turn, to reduce the chance of blindly overlapping with its partner. In contrast, the dashed green line shows the trajectory for the network that did not contain an attention schema. The average number of pixels selected by this network rose steeply from epoch 10 to the end of training at epoch 50. This rise suggests that the network was successfully learning to predict where its partner would place colored pixels. As a result, its best strategy was to color increasingly larger numbers of pixels which it could correctly place in a manner that avoided overlapping with its partner. The data therefore suggest that in the mismatched pair, the schema network became more interpretable and predictable to the non-schema network.

\subsection*{Discussion for Experiment 3}

\par The results of Experiment 3 show that having an attention schema confers a benefit on the joint coloring task. An agent needs to color as many pixels as it can on each trial, while avoiding overlap with the predicted coloring locations of its partner. Performance therefore depends on being able to predict where its partner will color on the current trial. Having an attention schema makes an agent better at this coordination. There are two possible reasons why performance might have improved. First, it could be that when a network has an attention schema, it is better able to predict the behavior of its partner. Second, it could be that when a network has an attention schema, it becomes more easily predictable to its partner. Both possibilities could be true simultaneously. The data suggest that the stronger effect is that when a network has an attention schema, it becomes more predictable to its partner. This result is consistent with the results of experiment 1, in which the largest impact of having an attention schema was to make a network more interpretable to other networks.

\par It is important to reiterate that the artificial mechanism of attention used here, a transformer mechanism, is very different from the biological mechanism in the human brain, and that the task used here, though well suited to testing the artificial agent, might be too simple to provide an insightful test of human participants. However, there are shared underlying principles, and therefore we feel it is valid to speculate about how the present results may reflect on the human case. One of the properties of the joint coloring task that makes it useful in the present context is that on each turn, an agent has a large range of options. It chooses how many and which pixels to color. In that sense, the task loosely resembles a human situation with open-ended choice. Should you reach for the coffee cup, the donut, the phone, or start a conversation? Moment by moment, the focus of your attention shifts to the feature of your world that will be the target of your next action. For this reason, being able to understand and predict attention is useful for understanding and predicting the behavior of others, and thus modeling the attention of others should be useful for cooperative interaction. In the joint coloring task, in a much simpler way, the same principles appear to be at work. Future studies may investigate a more complex version of the coloring task in humans.
\end{multicols}

\section*{General Discussion}
\begin{multicols}{2}
\par It has been proposed that the human attention schema, a model of attention, is important to our social cognition and our cooperative ability [1,6]. The goal of the present study was to test whether adding an attention schema to artificial agents using a transformer-style of attention would make them better cooperators in a joint task. We tested the transformer architecture partly because, although the mechanism is quite different from the biological attention mechanism in the human brain, some of the underlying principles of feature selection and influence on behavior are similar; and partly because we hoped to test the effects of attention schemas on the most common and widely used attention mechanism in machine learning. We used a version of an attention schema in which a network learns to predict its own attention state and uses those predictions to modify its attention state [18]. 

\par In Experiment 1, we found that adding an attention schema to an agent improved its ability to learn how to categorize the attentional state of another agent. Even more so, it improved the ability to have its own attentional state categorized by others. This finding is consistent with recent work suggesting that when an agent learns a predictive self-model, it also partly self-regularizes [19]. It learns to make itself more predictable. One interpretation of the present results, therefore, is that an agent with an attention schema, by virtue of learning to predict its own attentional states, regularizes its own attentional dynamics and makes itself more easily categorized, interpreted, and predicted by others. We speculate that this insight is generalizable – that in biological systems as well, one of the main adaptive advantages of self-models may be to make the self more regularized, more predictable, and thus a better target for the social cognition of others. In that speculation, having a self-model makes one more predictable and interpretable to others, thereby enhancing social cooperativity.

\par In Experiment 2, we found that the improvements caused by the presence of an attention schema were specific to a task involving the categorization of attention states. Adding an attention schema had no significant effect on other categorization tasks. This finding suggests that the effects are not the result of a general increase in network complexity. Instead, there is something about an attention schema that allows for improvement on attention-related tasks.

\par In Experiment 3, we tested agents in a simple joint task, a coloring task. Two agents took turns coloring in an image while minimizing overlap. On each turn, each agent chose pixels to color without knowing the pixels chosen by its partner on that turn. Ideal performance should require each agent to predict the choices of its partner, such that it can maximize the number of pixels it colors while minimizing overlap with its partner’s choices. The earned rewards were shared by both agents. We found, as hypothesized, that performance on the task was best when both agents contained an attention schema, weaker when only one of the two agents contained an attention schema, and weakest when neither agent in the pair contained an attention schema. Further analysis suggested that the largest effect of adding an attention schema to an agent may have been to make that agent more predictable to its partner, allowing the partner to perform better at the task. This may theoretically be achieved in any number of ways—making oneself more like one’s partner (thus, more predictable to the partner), making one’s moves simpler or less complicated, and so on.

\par The study of social cognition has focused almost exclusively on how mechanisms in the brain of person A can allow that person to better understand person B. However, it may be equally important, if not more so, to consider the mechanisms in the brain of person A that render that person more easily understood by person B. It is a matter of not just rendering oneself a better receiver of socially useful information, but also a better sender of socially useful information. We speculate that possessing an attention schema may provide both properties – a greater ability to understand others, and a greater ability to be understood by others – thereby improving social cooperation. In a competitive interaction, it is probably disadvantageous to be more easily understood by others. But in a cooperative interaction, rendering oneself more transparent, more interpretable, and ultimately more predictable, may be an advantage. We recognize that these suggestions emerge here out of a study of non-biological, artificial agents using artificial attention mechanisms, but we suggest that at least some of the underlying principles should be shared by all agents that use any form of attention, attention control, and attention schema.

\par The reason why an attention schema, in particular, may be of central importance, is that attention is almost entirely determinative of behavior. In the case of the artificial agents, attention determines the features that will most influence action. In the human case as well, people react to items that they are paying attention to, and do not react to items they are not attending. Human attention, the ability to control it, understand it, reveal it to others, and predict the attention of others, therefore plays a large role in social behavior. In a much-simplified way, the present study might be seen as reflecting the same kind of processes. 
\end{multicols}

\section*{Data availability}
All data, and code for running the experiments and analyzing the data are publicly available online at 
\begin{center}
  \url{https://github.com/kathrynfarrell/attention_schemas_in_anns}. 
\end{center}

\section*{Acknowledgements}
Supported by AE Studios grant 24400B1459FA010 and NIMH grant T32MH065214.
\\\\
\section*{References}
\begin{enumerate}[label={[\arabic*]}]
\item M. S. A. Graziano, S. Kastner, Human consciousness and its relationship to social  neuroscience: A novel hypothesis. Cognitive Neuroscience 2, 98-113 (2011).

\item T. W. Webb, M. S. A. Graziano, The attention schema theory: A mechanistic account of subjective awareness. Frontiers in Psychology, Vol 6, article 500, doi: 10.3389/fpsyg.2015.00500 (2015).

\item T. W. Webb, H. H. Kean, M. S. A. Graziano, Effects of awareness on the control of attention. Journal of Cognitive Neuroscience, 28: 842-851 (2016).

\item A. I. Wilterson, C. M. Kemper, N. Kim, T. W. Webb, A. M. W. Reblando, M. S. A. Graziano, Attention control and the attention schema theory of consciousness. Progress in Neurobiology, 195: 101844, doi: 10.1016/j.pneurobio.2020.101844 (2020). 

\item A. I. Wilterson, S. A. Nastase, B. J. Bio, A. Guterstam, M. S. A. Graziano, Attention, awareness, and the right temporoparietal junction. Proceedings of the National Academy of Sciences USA 118(25):e2026099118. doi: 10.1073/pnas.2026099118. PMID: 34161276 (2021).

\item M. S. A. Graziano, Consciousness and the Social Brain. New York: Oxford University Press (2013).

\item A. Pesquita, C. S. Chapman, J. T. Enns, Humans are sensitive to attention control when predicting others’ actions. Proceedings of the National Academy of Sciences USA 113, 8669-8674 (2016).

\item A. Guterstam, M. S. A. Graziano, Visual motion assists in social cognition. Proceedings of the National Academy of Sciences USA 117, 32165-32168 (2020).

\item A. Guterstam, H. H. Kean, T. W. Webb, F. S. Kean, M. S. A. Graziano, An implicit model of other people’s visual attention as an invisible, force-carrying beam projecting from the eyes. Proceedings of the National Academy of Sciences USA 116, 328-333 (2018).

\item A. Guterstam, M. S. A. Graziano, Implied motion as a possible mechanism for encoding other people’s attention. Progress in Neurobiology 190, article 101797 (2020).

\item A. Guterstam, A. I. Wilterson, D. Watchell, M. S. A. Graziano, Other people’s gaze encoded as implied motion in the human brain. Proceedings of the National Academy of Sciences USA 117, 13162-13167 (2020).

\item C. Renet, W. Randall, A. Guterstam, A motion aftereffect from viewing other people’s gaze. Frontiers in Human Neuroscience 18, doi 10.3389/fnhum.2024.1444428 (2024).

\item A. Vaswani, N. Shazeer, N. Parmar, J. Uszkoreit, L. Jones, A. N. Gomez, L. Kaiser, I. Polosukhin, Attention is all you need. arXiv: 1706.03762v5 (2017).

\item M. Zhao, D. Xu, T. Gao. From cognition to computation: A comparative review of human attention and transformer architectures. arXiv: 2407.01548 (2024).

\item J. van den Boogaard, J. Treur, M. Turpijn, A neurologically inspired network model for Graziano’s attention schema theory for consciousness. in International Work-Conference on the Interplay Between Natural and Artificial Computation: Natural and Artificial Computation for Biomedicine and Neuroscience. J. Ferrandez Vincente, J. Álvarez-Sánchez, F. de la Paz López, J. Toledo Moreo, H. Adeli, Eds. (Springer, Cham, Switzerland, 2017), pp. 10–21.

\item A. I. Wilterson, M. S. A. Graziano, The attention schema theory in a neural network agent: Controlling visuospatial attention using a descriptive model of attention. Proceedings of the National Academy of Sciences USA 118, e2102421118 (2021).

\item L. Piefke, A. Doerig, T. Kietzmann, S. Thorat, Computational characterization of the role of an attention schema in controlling visualspatial attention. arXiv: 2402.01056v2 (2024).

\item D. Liu, S. Bolotta, H. Zhu, Y. Bengio, G. Dumas, Attention Schema in Neural Agents. arXiv:2305.17375 (2023).

\item V. N. Premakumar, M. Vaiana, F. Pop, J. Rosenblatt, D. S. de Lucena, K. Ziman, M. S. A. Graziano, Unexpected Benefits of Self-Modeling in Neural Systems. arXiv preprint arXiv:2407.10188 (2024).

\item K. Ziman, S. C. Kimmel, K. T. Farrell, M. S. A. Graziano, Predicting the attention of others. Proceedings of the National Academy of Sciences USA 120, article e2307584120 (2023).

\item A. Paszke, S. Gross, F. Massa, A. Lerer, J. Bradbury, G. Chanan, et al., Pytorch: An imperative style, high-performance deep learning library. Advances in neural information processing systems, 32 (2019).

\item J. Howard, Imagenette, https://github.com/fastai/imagenette/ (accessed on Sept 1, 2024)

\end{enumerate}
\end{document}